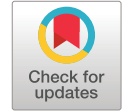

# Faster Metallic Surface Defect Detection Using Deep Learning with Channel Shuffling

**Siddiqui Muhammad Yasir[1] and Hyunsik Ahn[2,\*]**

[1]Department of Robot System Engineering, Tongmyong University, Busan, 48520, Korea
[2]School of Artificial Intelligence, Tongmyong University, Busan, 48520, Korea
*Corresponding Author: Hyunsik Ahn. Email: hsahn@tu.ac.kr
Received: 31 August 2022; Accepted: 24 October 2022

**Abstract:** Deep learning has been constantly improving in recent years, and a significant number of researchers have devoted themselves to the research of defect detection algorithms. Detection and recognition of small and complex targets is still a problem that needs to be solved. The authors of this research would like to present an improved defect detection model for detecting small and complex defect targets in steel surfaces. During steel strip production, mechanical forces and environmental factors cause surface defects of the steel strip. Therefore, the detection of such defects is key to the production of high-quality products. Moreover, surface defects of the steel strip cause great economic losses to the high-tech industry. So far, few studies have explored methods of identifying the defects, and most of the currently available algorithms are not sufficiently effective. Therefore, this study presents an improved real-time metallic surface defect detection model based on You Only Look Once (YOLOv5) specially designed for small networks. For the smaller features of the target, the conventional part is replaced with a depth-wise convolution and channel shuffle mechanism. Then assigning weights to Feature Pyramid Networks (FPN) output features and fusing them, increases feature propagation and the network's characterization ability. The experimental results reveal that the improved proposed model outperforms other comparable models in terms of accuracy and detection time. The precision of the proposed model achieved by mAP@0.5 is 77.5% on the Northeastern University, Dataset (NEU-DET) and 70.18% on the GC10-DET datasets.

## 1 Introduction

Computer vision methods are widely applied to detect surface defects on the metallic surface during the manufacturing process. Defects on metallic surfaces can arise as a result of numerous influencing factors in industrial production, such as material qualities and processing processes. These defects have an impact not just on the product's quality, but also on how it is used. As a result, finding such faults is crucial for quality assurance. Defect localization, recognition, and classification are commonly used image processing techniques for defect detection as the primary solution. Traditional image processing is widely utilized as a defect detection approach in combination with a manual visual inspection that is used to describe the defect with machine vision [1,2]. However, these systems require a significant number of complex threshold settings that are sensitive to changes in real-world environments and quickly affected by lighting, background, and other factors. As a result, ambient noise can readily impact these algorithms, resulting in a low-performing detection model. Additionally, they are unsuited for real-world deployments due to their limited flexibility [1]. The commonly used pre-defined features contain local binary patterns (LBP), a histogram of the oriented gradient (HOG), a gray-level co-occurrence matrix (GLCM), and other statistical features [3].

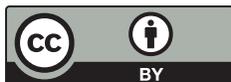

Deep learning-based techniques, convolutional neural networks (CNN) have been effectively implemented for metallic surface defect detection, especially with the rapid growth of deep learning [2]. The regional suggestion, regression operation, and generation of candidate defect boxes that are classified by CNNs are widely used in deep learning for defect detection [4]. Several researchers have presented metallic surface defect detection and classification by adding a feature extraction module to minimize processing time through CNN with a maximum pool [5,6]. In comparison to traditional methods based on image structure and statistical features combined with machine learning, CNN-based deep learning systems can automatically extract and recognize features in a network at the same time, eliminating the requirement for manual feature extraction [7].

Defect localization detecting bounding boxes or area segmentation allows the observer to find and understand surface flaws. Defect localization is essentially a subset of object detection. As a result, some studies regarded the detection of surface defects as a defect detection issue such as Faster R-CNN [6], Generic CNN Encoder [8], Single Shot Multibox Detector (SSD) [9], You Only Look Once (YOLO) [10]. In contrast to metallic surface flaws, where a large number of labeled images are not readily available, all of these models assume that a large number of labeled images are readily available. High-speed manufacturing requires automated feature extraction, different factors may involve in the real-time detection of metallic surface defect detection challenging, such as the high-speed manufacturing line, variety and large-scale changes in defects, random distribution, and non-defective interferences [1,4]. The purpose of this research is to reduce the amount of computation required for convolutional processes and to extract a feature map of metallic surface defects to balance speed and precision.

In this paper, a modified model based on YOLOv5 is proposed to solve the above-mentioned problems. The YOLO-v1 method is the base network for YOLO v2 and v3 added batch normalization, a high-resolution classifier, multi-scale training, and dimension clusters, while YOLO v4 and v5 fork from original Darknet, Cross Stage Partial (CSP) backbone and PA-NET neck, data augmentation, and auto-learning bounding box anchors. On the other hand, YOLOv5 largely relies on the backbone as a Darknet, which can result in a low confidence score, especially in a defect detection scenario. As a result, ShuffleNetv2 is implemented as a feature extractor, which reduces the time it takes to generate features at various levels. These features are useful in the second half of object detection networks while also improving the model's confidence score, precision, and detection speed. Thus, a modified variant effectively improved the detection precision of the model. The proposed model reached 77.5% mAP on NEU-DET and 71.10% mAP achieved on the GC10-DET dataset.

The main contributions of this study are:

1. Considering the high demands for detection speed in industrial production, a lightweight backbone network is proposed to reduce processing time and enhance overall model performance.
2. Due to the poor feature extraction performance of the currently existing deep convolution model, depth-wise convolution blocks were added to the model, which improved the model's confidence score.
3. Several experiments were carried out on two separate datasets, and the results demonstrated that the proposed model was robust and generalized.

The rest of the paper is organized as follows: Section 2 describes related work about defect detection in general. Section 3 describes in great detail the specific method of the paper, mainly describing the overall structure of the proposed method. Section 4 presents the details and the results of performing experiments on the datasets, which is followed by the conclusions drawn in Section 5.

## 2 Related Work

Over recent years, numerous kinds of research have been conducted for defect detection however they have not been confined to metallic surfaces. On the other hand, deep object detection approaches, such as Faster R-CNN [6], Generic CNN Encoder [8], SSD [9], and YOLO [10] are the primary methods for detecting defects.

Redmon et al. [11] introduced the YOLO-v1 model to reduce computing complexity to meet the real-time requirement of object detection and discovered that end-to-end real-time monitoring is possible. Meanwhile, the extraction of physical information and the building of structural features, when combined with machine learning, show significant advantages in the field of defect detection. As a result, deep learning is currently playing a critical role in defect detection. Many defect detection approaches based on deep object detection have been proposed. For example, mobile location, production monitoring, and surface and PCB defect detection.

Traditional image processing algorithms identify surface defects using manual features specified by specialists. Local Binary Pattern (LBP) [12], Histogram Oriented Gradient (HOG) [13], Grey-Level Co-occurrence Matrix (GLCM) [14], and other statistical features [15] are some of the most common manual features. Traditional image processing methods need complex threshold settings for defect recognition. If environmental factors change, the threshold values should be carefully adjusted. Otherwise, the lack of adaptability and robustness can make the algorithm unsuitable for the new environment. The deep learning method has numerous advantages in defect detection. The end-to-end detection method can be used to extract image features and reduce manpower and design time. One-stage and two-stage object detection algorithms are the most common deep learning techniques. Using a multilayer convolutional neural network, the one-stage approach collects image features (CNN). The default bounding box is set using the sliding window in the image characteristics and anchor, and the image features are fed into two network branches to classify the flaws. The selective search or regional proposed network (RPN) [16] is used in the first stage of the two-stage method to forecast the regional picture. The features of each region of interest (ROI) can then be retrieved by merging ROIs. For object classification and bounding box regression, ROI attributes are fed into two branches of the network. Object and defect detection models are improved by using algorithms like R-CNN [16], Fast and Faster R-CNN [6], SSD [9], and the YOLO series [17–19].

Several defect detection methods have been developed. Bui et al. proposed an improved knowledge-based neural network (KBNN) accompanying optimization algorithms and finite element analysis (FEA) to predict the spring-back angle of the metal sheets during bending [20]. Results showed that KBNN can accurately simulate the relationship between the rebound angle and process parameters. The problem of detecting complex defects in big structures is a difficult one. Ma et al. [21] provided a method for detecting intricate cluster defects that fitted the multi-scale dynamic variable node XFEM with a three-step detection scheme flawlessly. The model necessitates the deployment of additional sensors as well as the configuration of many parameters. Akhyar et al. [22] suggested a defect detection network based on RetinaNet focused to overcome the problem of tiny target and category imbalance. However, due to the model's anchor generation process and the addition of the Feature pyramid networks (FPN) structure, the model's computation amount is large, and the experiment's defect detection speed is only 0.08 s. As a result, it will be unable to meet the real-time detection demands.

To overcome the challenge of detecting materials under high-speed trains, Yao et al. [23] suggested a target detection model based on YOLOv3. The dense network takes the place of ResNet in the backbone and Spatial Pyramid Pooling (SPP) networks are employed to improve the model's detection precision. However, the dense network contains more layers and a larger feature map than the original model, resulting in a higher computed number of convolutions. As a result, the algorithm fails to strike a decent speed-accuracy balance and it is relatively difficult to deploy them in real-world industries. Thus, the defect detection area needs a smaller number of computations for convolutional processes and a reasonable feature map to balance the speed and precision of the model.

Although deep learning-based defect detection approaches have been extensively explored and considerably improved, there is still room for development in terms of defect detection accuracy and precision, as well as real-time rapid detection

## 3 Architectural Overview of YOLOv5 Detection Model

Object Detection based on CNN is largely used in object detection systems. Object detection with excellent performance is achieved using the YOLOv5 model. YOLO divides an image into grids, each of which identifies things inside its boundaries. They can be used to detect objects in real-time using data streams. A backbone will be used to pre-train the General Object Detector, and a head will be used to forecast classes and bounding boxes as shown in Fig. 1 Backbones can run on either GPU or CPU platforms. For Dense prediction, the Head (as shown in Fig. 1) can be one-stage e.g., YOLO [10,18,19,24], SSD [9], RetinaNet [25] or two-stage e.g., Faster R-CNN [6] for the sparse prediction object detector. Some layers (Neck) in recent object detectors collect feature maps and are located between the backbone and the head.

Cross Stage Partial CSPDarknet53 is used as a backbone and Spatial Pyramid Pooling (SPP) block in YOLOv4 to improve the receptive field, which separates the significant characteristics while maintaining network speed. The author created the Focus layer in the original version of the yolov5 backbone by using four slice operations in the higher structure of feature extraction. Every four neighboring pixels in a square, and produce a feature map with the number of channels on the focus layer. Equivalent to completing four down-sampling operations on the upper layer and then concatenating the results together. The model's function is to reduce parameters and accelerate the model without compromising the model's capacity to extract features.

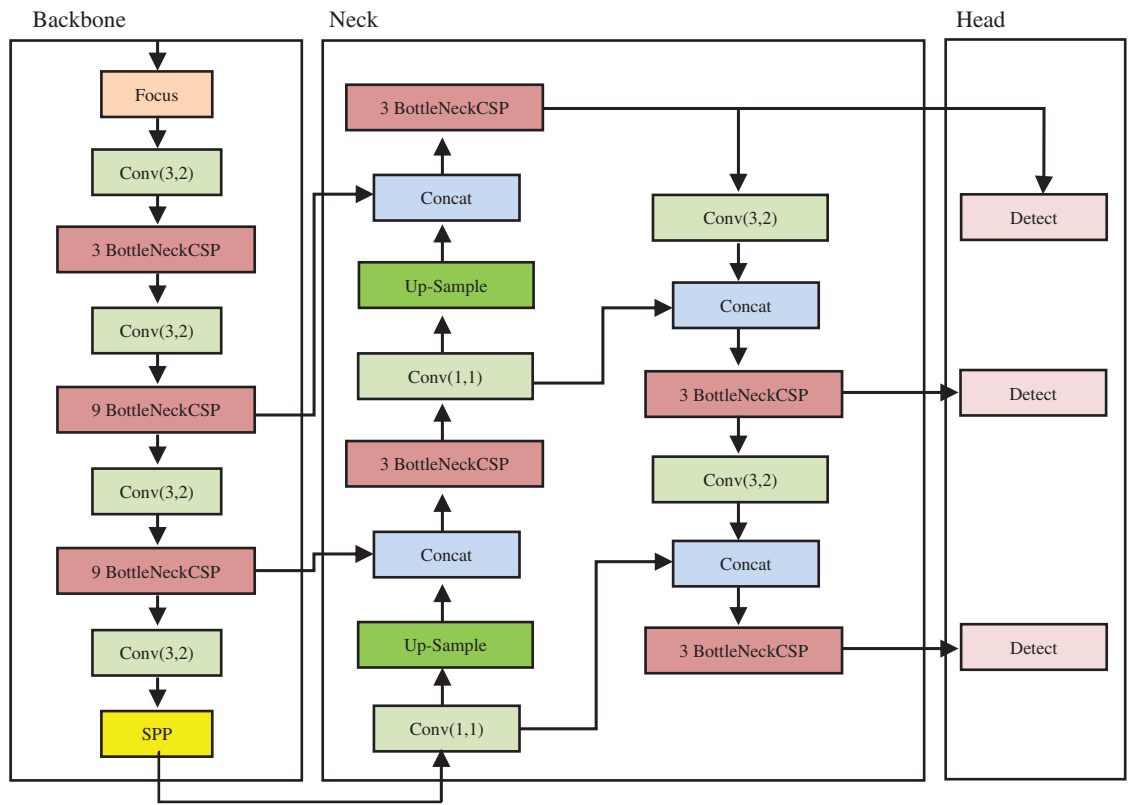

**Figure 1:** YOLOv5 default architecture (backbone, neck, and head), reference to the elements of this architecture modified in our work

A path aggregation network (PAN) is a tool for combining data from several backbone layers. The YOLOv3 (anchor-based) head is used for YOLOv4 [24]. Shortly after YOLOv4, YOLOv5 was released. This version is comparable to YOLOv4 in terms of performance and design. The key point of interest is that it is entirely developed in the PyTorch framework, rather than any form of the Darknet framework and that it is designed for accessibility and use in a wider range of development settings. Furthermore, the models in YOLOv5 are smaller, faster to train, and easier to use in real-world applications [26].

When the parameters are reduced, the Focus layer accelerates the model, and this advantage can only be obtained by using the GPU. In the take-and-go processing method, the GPU does not need to consider cache occupancy, and the focus layer is placed on the GPU device. Frequent slice operations will just clutter the cache and raise the computational processing load. Simultaneously, when the process is deployed, the conversion of the Focus layer is exceedingly difficult for fast detection.

### 3.1 Small Object Detection

Some effort has gone into developing systems that direct processing to specific areas of an input image, allowing us to change resolution and therefore overcome the constraint of having fewer pixels defining an object [27,28]. This technique, on the other hand, is better suited for non-time-sensitive systems that require multiple runs at different scales over a network. Nonetheless, the way someone approaches certain feature maps may be influenced by paying more attention to specific scales. Furthermore, changing the backbone, and looking at how feature maps can be treated may improve the detection much more accurately and faster. Feature pyramid networks (FPN) of many varieties can aggregate feature maps in different ways to improve a backbone in different ways. Such methods are rather effective [29].

Object detection can provide vital contextual information about the defect's surroundings and heavily inform the decision-making process inside defect detection. Smaller things translate to objects further away in this situation, giving the algorithm a more complete context to work with. These systems place a strong emphasis on inference time, at the expense of performance, if necessary, but work can be done to improve them at a low cost. Performance in this field is crucial, as even minor improvements can have a significant impact on the entire detecting system [29].

### 3.2 Light Weight Module Designs

A lightweight backbone structure is designed for metallic surface defect detection focused on YOLOv5 with NEU and CG10 datasets. The improved structure specially designed for small networks reduced the higher computed number of convolutions, which improves the accuracy and speed of detection. The backbone of YOLOv5 consists of convolutional blocks and bottleneck CSP. There are mainly three types of convolutional operations point wise, depth-wise, and group convolution in practice. Each conventional convolution layer is factorized by the baseline detection network into a depth-wise convolution layer and a point-wise convolution layer.

#### 3.2.1 Point-Wise Convolution

The point-wise convention uses a fixed $1 \times 1$ kernel size that iterates every single point, a standard convolution mainly used to collect data from different channels. The classic convolution convolved the depth of different channels as an input feature map in both spatial and channel dimensions. It is used with depth-wise convolutions to produce a class of efficient convolutions known as depth-wise separable convolutions. The depth-wise convolution could convolve the input feature map in the spatial dimension, but it eliminates information flow between the multiple channels. As a result, the output size remains constant, and it plays a role in controlling the number of channels. It is most commonly used for dimensional reduction to reduce the number of channels, which is critical because it can greatly reduce the amount of computation.

#### 3.2.2 Group Convolution (GConv)

This sort of convolution was first developed by the AlexNet [30] architecture, and it has lately gained popularity due to its use in ResNeXt [31]. By applying convolutional filters to all input feature maps, standard convolutional layers generate output feature maps resulting in high computational costs. Group convolution, on the other hand, lowers the computational cost by separating the input features into mutually exclusive groups, each of which creates its outputs. According to the proposed approach, the group size should be carefully selected based on the target platform and task. For a modest task like defect detection, adopting a high number of group convolutional layers isn't appropriate because adding more channels can reduce accuracy by significantly increasing computing costs. With numerous group convolutional layers and the channel shuffle operation, a more powerful architecture is proposed in this research.

#### 3.2.3 Depth-Wise Convolution

As an efficient alternative to the standard convolution process, a Depth-wise separable convolution layer has been developed. An efficient kind of Neural Network (NN) known as MobileNet [32] is created by replacing a typical 3-D convolution with a 2-D depth-wise convolution followed by a 1-D point-wise convolution. To improve the accuracy of compact models, the proposed model uses depth-wise convolutions on shuffled channels combined with group-wise 1 × 1 convolutions. ShuffleNet [33], and MobileNetv2 [32] boosted efficiency even more by introducing shortcut connections that assist deep network convergence. Overall, several efficient neural network topologies have been presented, which can be used when constructing a NN model, especially for metallic surface defect detection.

## 4 Proposed Architecture

In YOLOv5, the modified ShuffleNetv2 [34] is replaced with the present backbone. The backbone element is responsible for extracting feature maps from the input image. This is an important stage in any object detector because it is the principal structure in charge of gathering contextual information from the input image and abstracting that information into patterns.

ShuffleNetv2 is a convolutional neural network that is optimized for speed rather than indirect metrics such as FLOPs. DarkNet [35] makes use of such connections to keep as much data as possible safe as it travels over the network. Putting these structures in place necessitates breaking them down into their essential components and ensuring that the layers interact properly. This involves ensuring that the feature map dimensions are correct, which may necessitate significantly changing the scaling factor for the model's width and depth. Tasks generally seek maximum accuracy while working within a limited computing budget set by the target platform. This encourages a number of projects, such as Xception [36], MobileNet [37], ShuffleNet [34], CondenseNet [38] to create light-weight architectures with improved speed-accuracy tradeoffs. These works rely heavily on group convolution and depth-wise convolution. Putting these structures in place necessitates breaking them down into their basic components and ensuring that the levels interact properly. This involves maintaining the correct feature map size, which may involve slightly adjusting the scaling factor for the model's width and depth.

### 4.1 Proposed Backbone

The improved network architecture in Fig. 2 depicts the enhanced network design. The specially designed depth-wise structure is added to the core network architecture to enhance feature dissemination and reuse, based on the original ShuffleNetv2 architecture. To ensure efficient information communication, two 2D convolution blocks to our model are embedded. Each layer's output was coupled to all the following layers in each 2D convolution block. The design principle of the bottleneck unit in Fig. 2 is using the channel shuffle function, which is specifically built for small networks functioning as a residual block. The 3 × 3 layer is added to the residual branch, followed by computationally efficient 3 × 3 depth-wise convolution on the bottleneck feature map. The initial 1 × 1 layer is then replaced by a point-wise group convolution followed by a channel shuffle operation.

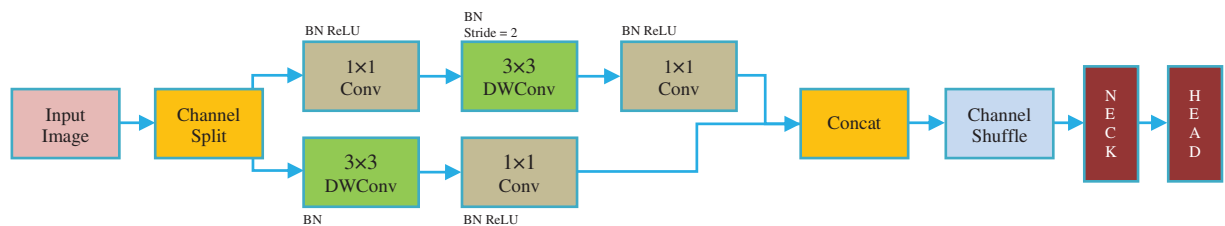

**Figure 2:** Building blocks of the proposed backbone structure

Since each building block is so efficient, more feature channels and network capacity may be used. The fundamental issue with lightweight networks is that only a limited number of feature channels are affordable within a certain processing budget (FLOPs). In [30], two strategies are used to expand the number of channels without considerably raising FLOPs: point-wise group convolutions and bottleneck-like structures. Then, to enable information communication between different sets of channels and increase accuracy, a "channel shuffle" extremely efficient process is adopted.

On the computational side, where resources are scarce, the ShuffleNetv2 design concept has several reference meanings. C3 Layer is an upgraded version of the CSPBottleneck developed by the YOLOv5 authors that are simpler, faster, lighter, and delivers better outcomes with roughly comparable losses. C3 Layer, on the other hand, employs multiplexed convolution. Tests have shown that using C3 Layer and C3 Layer with a higher number of channels frequently consumes more cache space and reduces running speed. This module is intended for ImageNet ranking. When there aren't as many different types of realistic real-time scenarios. Backbone architecture similar to ShuffleNetv2 is used to replace 1024 convolution and $5 \times 5$ pooling. There is an improvement in accuracy and a significant gain in terms of speed.

### 4.2 Influence of the Backbone

The YOLOv5's default backbone not only appears to degrade performance in most circumstances, but its inference time is also substantially longer, providing little reason to examine it further at this point. We conclude that the proposed model is a better fit for small-scale object detection in general. This may be owing to the smaller size models however not having a network deep enough to appreciate the benefits of a DarkNet backbone, whereas the proposed model does a decent job of conserving the detailed feature map.

## 5 Experiments

Several different experiments were carried out in this section to evaluate the suggested method utilizing real flaws in metallic surfaces. The setup parameters for the experiment were as follows: AMD Ryzen 7 6000 CPU, 64 GB DDR4 RAM, NVidia GTX2080Ti GPU with 24 GB memory, and the Ubuntu 18.04 LTS operating system. The results of the experiment are then explained in both visual and quantitative formats. There are two different datasets Table 1 is used in this research and presented the results in detail.

**Table 1:** Comparison of NEU-DET and GC10-DET datasets

| Dataset | Scale | Type number | Defect types |
|---------|-------|-------------|--------------|
| NEU-DET | 1800 | 6 | Rolled-in scale, patches, crazing, inclusion, pitted surface, scratches |
| CG10-DET | 3570 | 10 | Punching, weld line, crescent gap, inclusion, water spot, oil spot, silk spot, rolled pit, crease, waist folding |

### 5.1 Description of NEU-DET and GC10-DET Dataset

NEU-DET is a surface defect dataset created by Northeastern University (NEU) that covers six different types of surface defects. There are 1800 gray-scale defects are found on the hot-rolled steel strip's surface in the collection, with 300 samples in each class of surface defects [12,39].

The GC10-DET dataset has significant difficulties in terms of fault categories, image count, and data scale. Additionally, standard detection methods are inefficient and inaccurate in the complicated real-world context. The surface defect dataset GC10-DET was obtained in an actual industry [39]. There are 3570 gray-scale images in the dataset. The specific defect classes included in the GC10- DET dataset are Punching, Welding line, Crescent gap, Water spot, Oil spot, Silk spot, Inclusion, Rolled pit, Crease, and Waist folding.

### 5.2 Performance Evaluation

For performance evaluation, recall and mean average precision (mAP) is used. For each fault category, recall is the ratio of successfully recognized images to all testing images. The average detected precision for each defect category is represented by the mean average detected precision for all defect categories is represented as mAP.

The proposed model was compared with several state-of-the-art methods, including SSD, YOLOv4, YOLOv5, and Faster-RCNN [6,9,24,26]. To be comprehensive, the optimal parameter tuning is presented in this section. The above-mentioned deep methods adopt a pre-trained model on the ImageNet, which can be helpful to extract basic image features including edge, texture, and so on. Therefore, the SSD method utilizes VGG16 as the pre-trained model, Darknet19 model, YOLOv4 uses Darknet53 model, and Faster R-CNN adopts the Resnet50 model and the parameter tuning for these models respectively explained as follows:

The learning rate is determined by training experience; a higher learning rate indicates higher weight updates, which can speed up model convergence; however, if the learning rate is too high, the training may fluctuate and may lead to slow convergence of the training process. The initial learning rate is set from 0.1 to 0.00001, and the best one is selected through experiments. Thus, the best learning rate was obtained as follows: SSD (0.0005), Faster-RCNN (0.01), and YOLOv5 (0.0005).

The weight decay is a coefficient of the regular term in the loss function. Thus, the setting of weight decay depends on the loss function used to alleviate over-fitting. According to the loss function and experiments, the final settings of weight decay are as follows: SSD (0.00005), Faster-RCNN (0.0001), and YOLOv5 (0.00005).

## 6 Results and Discussion

The results of the metallic steel surface defect detection are shown in Figs. 3 and 4. Steel strip defects occur in a variety of sizes and shapes, with a wide range of properties. The "silk spot" defect is enormous, but its characteristics are abstract due to the size discrepancies in "oil stains" and other forms of defects. Sections 6.1 and 6.2 explain detailed comparisons of both datasets and each class. Furthermore, the efficacy of the proposed and traditional five defect detection models as images of single and testing sets of defected steel sheets is compared in Table 6.

### 6.1 Comparison of Accuracy and Recall on NEU-DET Dataset

The detailed comparison of Recall on the NEU-DET dataset is shown in Table 2 and the detection results of NEU-DET are shown in Fig. 3 visually. The proposed method can obtain the best results on the defects of the NEU-DET dataset, while the SSD has a slightly higher Recall than the proposed method. Tables 2, and 3 demonstrate the detailed comparison results about mean average precision and recall against each class.

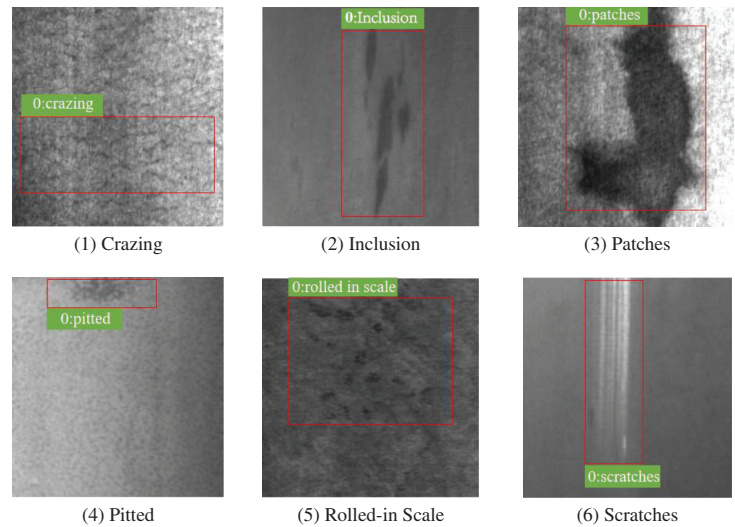

**Figure 3:** Results of detection on the NEU-DET dataset. The images are arranged in the following order: (1) Crazing, (2) Inclusion, (3) Patches, (4) Pitted surface, (5) Rolled in scale, and (6) Scratches

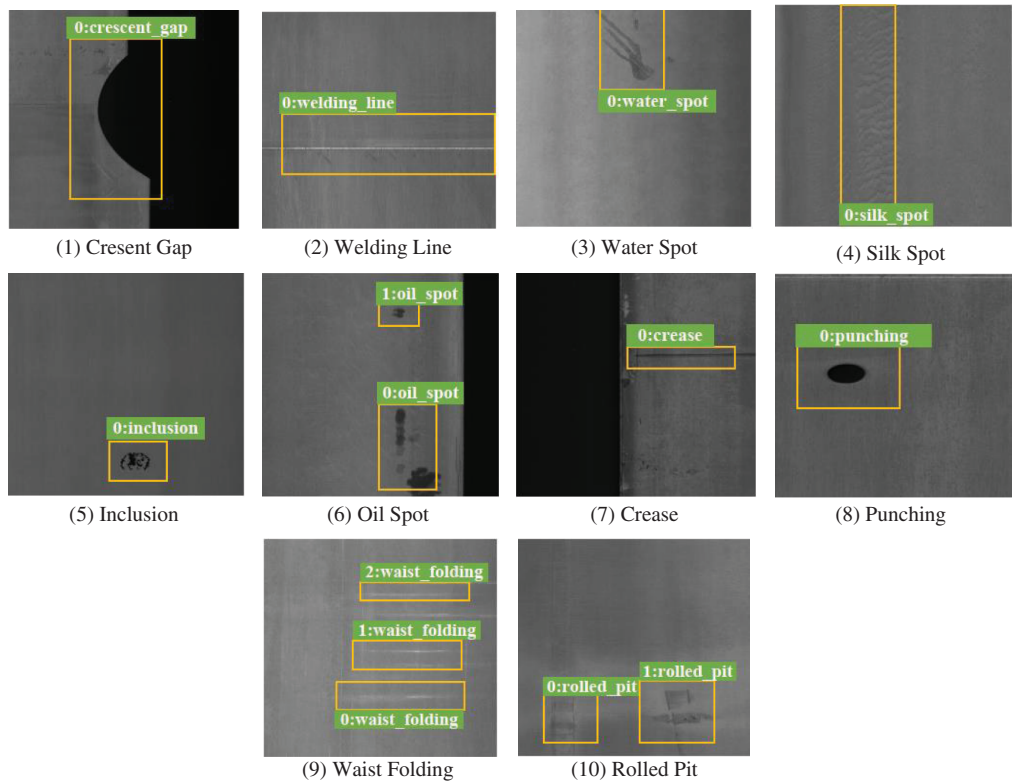

**Figure 4:** Results of detection on the GC10-DET dataset. The images are arranged in the following order: (1) Crescent gap, (2) Welding line, (3) Water spot, (4) Silk spot, (5) Inclusion, (6) Oil spot, (7) Crease, (8) Punching, (9) Waist folding, and (10) Rolled pit

**Table 2:** Recall on the NEU-DET dataset is compared using five distinct models

| Defect types | Recall | | | | |
|---|---|---|---|---|---|
| | SSD | Faster-RCNN | YOLOv5 | YOLOv6 | Proposed model |
| Crazing | 0.665 | 0.474 | 0.692 | 0.781 | 0.631 |
| Inclusion | 0.974 | 0.903 | 0.655 | 0.889 | 0.768 |
| Pitted surface | 0.875 | 0.851 | 0.561 | 0.671 | 0.624 |
| Patches | 0.871 | 0.843 | 0.923 | 0.812 | 0.622 |
| Roll in scale | 0.632 | 0.581 | 0.602 | 0.710 | 0.572 |
| Scratches | 0.790 | 0.771 | 0.716 | 0.805 | 0.690 |

**Table 3:** Mean average precision (mAP) on the NEU-DET dataset is compared using five distinct models

| Defect types | Precision | | | | |
|---|---|---|---|---|---|
| | SSD | Faster-RCNN | YOLOv5 | YOLOv6 | Proposed model |
| Crazing | 0.411 | 0.374 | 0.221 | 0.552 | 0.591 |
| Inclusion | 0.796 | 0.784 | 0.580 | 0.781 | 0.867 |
| Patches | 0.839 | 0.853 | 0.872 | 0.872 | 0.951 |
| Pitted surface | 0.839 | 0.795 | 0.539 | 0.739 | 0.859 |
| Roll in scale | 0.621 | 0.545 | 0.335 | 0.601 | 0.656 |
| Scratches | 0.836 | 0.792 | 0.750 | 0.838 | 0.726 |
| **mAP** | **72.36** | **69.05** | **54.95** | **73.05** | **77.50** |

The six categories of metal surface defects are difficult to discern using the YOLO models, as demonstrated in Tables 2 and 3. The reason for this could be those surface imperfections are typically tiny in size. Thus, the default YOLOv5 with fixed scale detection cannot effectively tackle this. On the other hand, the proposed model uses multi-scale cells to better distinguish multi-scale faults with 0.775 mAP. Even though Faster-RCNN uses anchor boxes to solve this problem, it still performs worse than the proposed model. The proposed method performs the highest mean average precession values except in particular one defect type "Crazing".

### 6.2  Comparison of Accuracy on GC10-DET Dataset

The detailed comparative findings of Recall on the GC10-DET dataset are shown in Tables 4, 5, and Fig. 4 depict several GC10-DET detection results. The suggested technique performs best on defects of GC10-DET dataset whereas SSD performs slightly better Recall on defect type "Rolled pit" and Faster-RCNN performs slightly better on defect type "Crease".

The comprehensive comparative findings of Recall on the GC10-DET dataset are presented. The proposed technique achieves the greatest results on the faults. The SSD, on the other hand, is slightly greater on defect types such as "Welding Line", "Rolled Pit", and "Waist Folding". On defect type "Punching", faster-RCNN outperforms all the others.

**Table 4:** Comparison of recall on GC10-DET dataset, the proposed model performs the highest

| Defect types | Recall | | | | |
|---|---|---|---|---|---|
| | SSD | Faster-RCNN | YOLOv5 | YOLOv6 | Proposed model |
| Punching | 0.964 | 0.964 | 0.964 | 0.804 | 0.675 |
| Welding line | 1.000 | 0.623 | 0.899 | 0.785 | 0.667 |
| Crescent gap | 0.968 | 0.968 | 0.871 | 0.745 | 0.569 |
| Water spot | 0.696 | 0.696 | 0.609 | 0.698 | 0.519 |
| Oil spot | 0.848 | 0.761 | 0.565 | 0.806 | 0.691 |
| Silk spot | 0.956 | 0.708 | 0.542 | 0.924 | 0.588 |
| Inclusion | 0.578 | 0.551 | 0.311 | 0.571 | 0.367 |
| Rolled pit | 0.667 | 0.333 | 0.333 | 0.543 | 0.333 |
| Crease | 0.571 | 1.000 | 0.429 | 0.629 | 0.467 |
| Waist folding | 0.999 | 0.900 | 0.700 | 0.849 | 0.690 |

**Table 5:** Comparison of mean average precision (mAP) on GC10-DET dataset, the proposed model performs the highest

| Defect types | Precision | | | | |
|---|---|---|---|---|---|
| | SSD | Faster-RCNN | YOLOv5 | YOLOv6 | Proposed model |
| Punching | 0.860 | 0.899 | 0.836 | 0.752 | 0.895 |
| Welding line | 0.974 | 0.554 | 0.841 | 0.813 | 0.885 |
| Crescent gap | 0.861 | 0.872 | 0.752 | 0.772 | 0.948 |
| Water spot | 0.552 | 0.599 | 0.495 | 0.698 | 0.654 |
| Oil spot | 0.612 | 0.653 | 0.529 | 0.589 | 0.622 |
| Silk spot | 0.689 | 0.579 | 0.325 | 0.742 | 0.750 |
| Inclusion | 0.168 | 0.194 | 0.536 | 0.165 | 0.351 |
| Rolled pit | 0.105 | 0.364 | 0.236 | 0.381 | 0.367 |
| Crease | 0.527 | 0.736 | 0.629 | 0.831 | 0.621 |
| Waist folding | 0.99 | 0.818 | 0.554 | 0.915 | 0.925 |
| **mAP** | **63.38** | **62.68** | **57.33** | **66.58** | **70.18** |

As shown in Tables 3 and 5, the default YOLO backbones struggle to differentiate between the six types of defects in the NEU-DET dataset and the ten types of defects in the CG10-DET dataset. The explanation for this could be that the defects on the surface are often on a tiny scale, which cannot be treated well by default YOLOv5 with fixed scale detection. The proposed technique, on the other hand, distinguishes multi-scale defects better, and the mAP can reach 0.7750 and 0.7018 respectively. While the other two models are lower than the proposed technique.

**Table 6:** Comparison of computational time on two datasets for the traditional model

| Dataset | Test type | Method | | | | |
|---------|-----------|--------|--------------|---------|---------|----------------|
| | | SSD | Faster R-CNN | YOLOv5 | YOLOv6 | Proposed model |
| NEU-DET | Testing set | 51 ms | 47 ms | 54.75 ms | 34.85 ms | 21 ms |
| | Single image | 7 ms | 7 ms | 8.46 ms | 5.15 ms | 3 ms |
| CG10-DET | Testing set | 29 ms | 63 ms | 46.3 ms | 41.6 ms | 24 ms |
| | single image | 5 ms | 11 ms | 8.67 ms | 7.67 ms | 4 ms |

### 6.3 Computational Time Comparison

The proposed model is capable of completing tasks in a reasonable amount of time. To process one NEU-DET image, the proposed model took the almost same amount of time as SSD *vs.* the proposed model, 7 *vs.* 3 ms, for a total time of 51 *vs.* 21 ms for the entire testing set. On the GC10-DET, the proposed model obtained a second computing speed of 29 *vs.* 24 ms, despite the results for the single image test being the same at 5 and 4 ms.

## 7 Conclusion

In this paper, a deep learning model for metallic surface defect detection based on modified YOLOv5 was proposed. The improved deep learning model was suggested to enhance the performance for small sizes and complex shapes of defects and to reduce computational time. To improve the network's ability to extract features, depth-wise convolution was utilized to reconstruct the feature extraction network. The detection effect of the improved proposed model was better than other models when tested on the NEU-DET dataset at 77.5% and CG10-DET dataset at 70.18% mAP. Simultaneously, the proposed model's dominance was demonstrated by comparing it to other widely used target defect algorithms. To achieve maximum accuracy, a backbone with depth-wise convolution and data augmentation was improved. On the other hand, the default backbone of YOLOv5 was heavily reliant on the backbone as a Darknet, which can result in a poor confidence score, particularly in the defect detection scenario. As a result, the modified model with a backbone based on ShuffleNetv2 as a feature extractor was introduced, which minimizes the time required to generate features at various levels helpful in the second half of object detection networks while simultaneously boosting the model's confidence score, precision, and detection speed.

Finally, extensive experiments demonstrated that the proposed approach is reliable for detecting metallic surface defects. However, steel plates moving with great speed and extremely minute faults even in high-resolution images cannot be recognized successfully. In future work, the proposed research will be extended to enhance the performance, especially small-size defect detection.

**Funding Statement:** This research was supported by a grant from Tongmyong University Innovated University Research Park (I-URP) funded by Busan Metropolitan City, Republic of Korea.

**Conflicts of Interest:** The authors declare that they have no conflicts of interest to report regarding the present study.